\documentclass[runningheads]{llncs}
\pdfoutput=1
\usepackage{graphicx}

\usepackage{tikz}
\usepackage{comment}
\usepackage{amsmath,amssymb} 
\usepackage{color}

\usepackage[pagebackref=true,breaklinks=true,colorlinks,citecolor=cyan,linkcolor=red,bookmarks=false]{hyperref}
\usepackage[misc]{ifsym}

\usepackage{multirow}
\usepackage{colortbl}
\usepackage{caption}
\usepackage{subcaption}
\usepackage{graphicx}

\usepackage{booktabs}
\usepackage{multirow}
\usepackage{makecell}
\usepackage{bbding}
\usepackage[misc]{ifsym}

\begin{document}

\pagestyle{headings}
\mainmatter
\def\ECCVSubNumber{5207}  

\title{MorphMLP: An Efficient MLP-Like Backbone for Spatial-Temporal Representation Learning} 



\titlerunning{MorphMLP}
%

\author{
    David Junhao Zhang\textsuperscript{\rm 1}\thanks{Contribute equally. \dag Work is done during internship at Meitu, Inc.}\textsuperscript{\rm $\dag$},
    Kunchang Li\textsuperscript{\rm 3,4$\star$},
    Yali Wang\textsuperscript{\rm 3$\star$},
    Yunpeng Chen\textsuperscript{\rm 2},
    Shashwat Chandra\textsuperscript{\rm 1},
    Yu Qiao\textsuperscript{\rm 3,5},
    Luoqi Liu\textsuperscript{\rm 2},
    Mike Zheng Shou\textsuperscript{\rm 1\Letter}
}
\authorrunning{David J. Zhang et al.}
%
\institute{
    $^1$National University of Singapore \quad $^2$Meitu, Inc\\
    $^3$ShenZhen Key Lab of Computer Vision and Pattern Recognition, SIAT-SenseTime Joint Lab, Shenzhen Institutes of Advanced Technology, Chinese Academy of Sciences\\
    $^4$University of Chinese Academy of Sciences \quad  $^5$Shanghai AI Laboratory\\
}

\maketitle
\begin{abstract}

Recently,
MLP-Like networks have been revived for image recognition.
However, 
whether it is possible to build a generic MLP-Like architecture on video domain has not been explored,
due to complex spatial-temporal modeling with large computation burden.
To fill this gap,
we present an efficient self-attention free backbone, 
namely MorphMLP, 
which flexibly leverages the concise Fully-Connected (FC) layer for video representation learning.
Specifically,
a MorphMLP block consists of two key layers in sequence,
i.e.,
$\tt{MorphFC_s}$ and $\tt{MorphFC_t}$,
for spatial and temporal modeling respectively.  
$\tt{MorphFC_s}$ can effectively capture core semantics in each frame,
by progressive token interaction along both height and width dimensions.
Alternatively,
$\tt{MorphFC}_t$ can adaptively learn long-term dependency over frames,
by temporal token aggregation on each spatial location. 
With such multi-dimension and multi-scale factorization,
our MorphMLP block can achieve a great accuracy-computation balance.
Finally,
we evaluate our MorphMLP on a number of popular video benchmarks.
Compared with the recent state-of-the-art models,
MorphMLP significantly reduces computation but with better accuracy,
e.g.,  
MorphMLP-S only uses 50\% GFLOPs of VideoSwin-T but achieves 0.9\% top-1 improvement on Kinetics400,
under ImageNet1K pretraining.
MorphMLP-B only uses 43\% GFLOPs of MViT-B but achieves 2.4\% top-1 improvement on SSV2,
even though MorphMLP-B is pretrained on ImageNet1K while MViT-B is pretrained on Kinetics400. Moreover, our method adapted to the image domain
outperforms previous SOTA MLP-Like architectures. Code is available at \url{https://github.com/MTLab/MorphMLP}.

\keywords{MLP, Video and Image Recognition, Representation Learning}

\end{abstract}
\section{Introduction}
\label{sec:intro}

Since the seminal work of Vision Transformer (ViT) \cite{vit}, 
attention-based architectures have shown the great power in a variety of computer vision tasks, ranging from image domain \cite{swin,cswin,pvt,t2t} to video domain \cite{timesformer,motionformer,video_swin,video_Transformer,vidtr}.
However,
recent studies have demonstrated that,
self-attention maybe not critical and it can be replaced by simple Multiple Layer Perceptron (MLP) \cite{mixer}.
Following this line,
a number of MLP-Like architectures have been developed on image-domain tasks with promising results \cite{vip,cyclemlp,gmlp,smlp,resmlp,mixer}.

\begin{figure*}[t]
    \centering
        \begin{minipage}[t]{0.38\linewidth}
        \centering
        \includegraphics[width=0.9\textwidth]{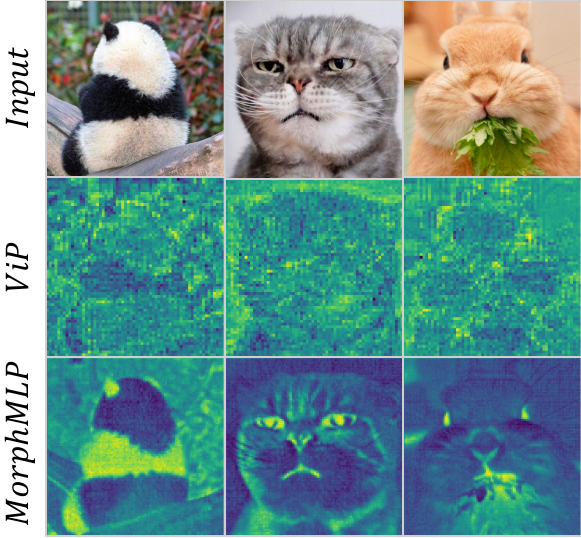}
 
        \caption{Visualization of spatial feature in 3rd layer.}
        \label{vis}
    \end{minipage}
    \hspace{2mm}
    \begin{minipage}[t]{0.58\linewidth}
        \centering
        \includegraphics[width=0.9\textwidth]{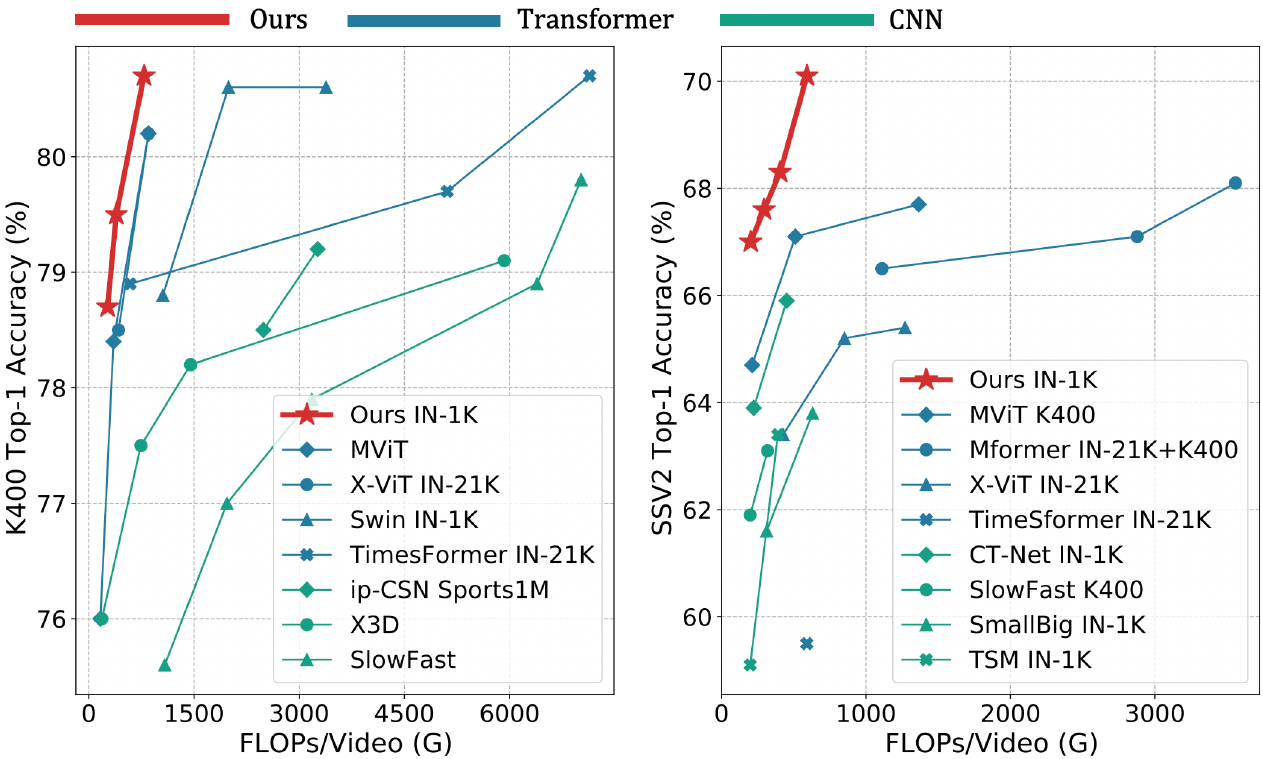}
        \caption{Our MorphMLP vs. other SOTA Transformers and CNNs for video classification.
        Left: Kinetics400 \cite{k400}; Right: SthV2 \cite{sth}.}
        \label{fig:intro}
    \end{minipage}
  
\end{figure*}

A natural question is that,
is it possible to design a generic MLP-Like architecture for video domain?
Unfortunately,
it has not been explored in the literature,
to our best knowledge.
Motivated by this fact,
we analyze the main challenges of using MLP on spatial-temporal representation learning.
\textbf{First},
from the \textit{spatial} perspective,
we find that the current MLP-Like models lack progressive understanding of semantic details.
This is mainly because that,
they often operate MLP globally on all the tokens in the space,
while ignoring hierarchical learning of visual representation.
For illustration,
we visualize the feature map of the well-known MLP-like model (i.e., ViP \cite{vip}) in Fig.\ref{vis}.
Clearly,
it suffers from difficulty in capturing key details, 
even in the shallow layer.
Hence,
how to discover semantics in each frame is important for designing spatial operation of MLP-like video backbone.
\textbf{Second},
from the \textit{temporal} perspective,
the critical challenge is to learn long-range dependencies over frames.
As shown in Fig. \ref{fig:intro},
the current video-based transformers can leverage self-attention to achieve this goal,
but with huge computation cost.
Hence,
how to efficiently replace self-attention for long-range aggregation is important for designing temporal operation of MLP-like video backbone.

\begin{figure*}[t]
\begin{center}
\includegraphics[width=0.98\columnwidth]{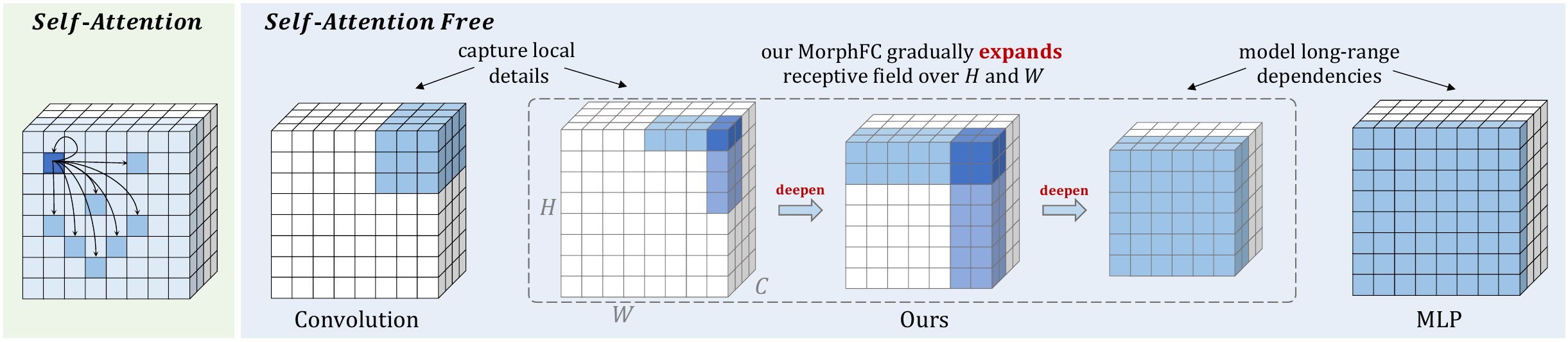}

\end{center}
\caption{
Overview of progressive token construction in MorphMLP. 
}

\label{fig:res}
\end{figure*}

To tackles these challenges,
we propose an effective and efficient MLP-like architecture,
namely MorphMLP,
for video representation learning.
Specifically,
it consists of two key layers,
i.e.,
$\tt MorphFC_s$ and $\tt MorphFC_t$, 
which leverage the concise FC operations on spatial and temporal modeling respectively. Our $\tt MorphFC_s$ can effectively capture core semantics in the space,
as shown in Fig. \ref{vis}.
The main reason is that,
we gradually expand the receptive field of visual tokens along both height and width dimensions as shown in Fig. \ref{fig:res}.
Such progressive token design brings two advantages in spatial modeling,
compared with the existing MLP-like models, e.g., ViP \cite{vip}.
First,
it can learn hierarchical token interactions to discover the discriminative details,
by operating FC from small to big spatial regions.
Second,
such small-to-big token construction can effectively reduce computation of FC operation for spatial modeling.



 
Moreover,
our $\tt MorphFC_t$ can adaptively capture long-range dependencies over frames.
Instead of exhausting token comparison in self-attention,
we concatenate the features of each spatial location across all frames into a temporal chunk. 
In this way, 
each temporal chunk can be processed efficiently by FC,
which adaptively aggregates token relations in the chunk to model temporal dependencies.
Finally,
we build up a MorphMLP block by arranging $\tt MorphFC_s$ and $\tt MorphFC_t$ in sequence,
and stack these blocks into our generic MorphMLP backbone for video modeling.
On one hand,
such hierarchical manner can enlarge the cooperative power of $\tt MorphFC_s$ and $\tt MorphFC_t$ to learn complex spatial-temporal interactions in videos.
On the other hand,
such multi-scale and multi-dimension factorization allows our MorphMLP to achieve a preferable balance between accuracy and efficiency.
 
To our best knowledge, 
we are the first to build efficient MLP-Like architecture for video domain.
Compared with the recent state-of-the-art video models,
MorphMLP significantly reduces computation but with better accuracy.

We further apply our method to an image classification task on ImageNet-1K\cite{imagenet}and a semantic segmentation task on ADE20K\cite{ade20k}, by simply removing the temporal dimension of the video. Our method adapted to the image domain achieves competitive results compared to previous SOTA MLP-Like architectures.

\section{Related Work}
\label{sec:related}

\noindent
\textbf{Self-Attention based backbones.}
Vision Transformer (ViT) \cite{vit} firstly applies Transformer architecture to a sequence of image tokens.
It utilizes multi-head self-attention to capture long-range dependencies,
thus achieving surprising results on image classification. 
 Following works\cite{swin,visualtransformer,cswin,t2t,pvt,botnet} make a series of breakthroughs to achieve state-of-art performance on several image tasks, i.e., semantic segmentation \cite{segformer,TrSeg} and object detection \cite{detr,d_detr}. In video domain, a couple of woks \cite{motionformer,timesformer,vidtr,focal,video_swin,mvit} explore space-time self-attention to model spatial-temporal relation and achieve  state-of-the-art performance.
It seems that self-attention based architectures have been gradually dominating the computer vision community.

In this paper,
we aim to explore a simple yet effective self-attention free architecture, which builds upon the FC layer to extract features.  Our comparisons show that  MorphMLP can achieve competitive results compared with Transformers not only in images but also in videos without self-attention layers.

\noindent
\textbf{CNN based backbones.} CNNs \cite{senet,mobilenetv1,mobilenetv2,regnet,resnext,li2020hard,li2021else} have dominated vision tasks in the past few years. 
In image domain, beginning with AlexNet\cite{alexnet}, more effective and deeper networks, 
VGG\cite{vgg}, GoogleNet\cite{google}, ResNet\cite{resnet}, DenseNet\cite{densenet} and EfficentNet\cite{efficientnet} are proposed and achieve great success in computer vision.  In the video domain, several works\cite{c3d,i3d,r(2+1)d,slowfast} explore how to utilize convolution to learn effective spatial-temporal representation. 
However, the typical spatial and temporal convolution are so local that they struggle to capture long-range information well even if stacked deeper. 
A series of works propose efficient modules
(e.g., Non-local\cite{non_local}, Double Attention\cite{double_attention}) to  enhance local features via integrating long-range relation. The improvement of these methods can not be achieved without the supplement of self-attention layers.

In contrast, we propose the MorphMLP, which is self-attention free but not limited to capture local structure. The FC filter of $\tt{MorphFC}$ operates from small to big spatial regions. Meanwhile, the $\tt{MorphFC_t}$ can capture long-term temporal information.


\noindent\textbf{MLP-Like based backbones.}
Recent works\cite{resmlp,gmlp,mixer,smlp} try to  replace self-attention layer with FC layer to explore the necessity of self-attention in Transformer architecture. But they suffer from dense parameters and computation. 
\cite{vip,hire,sparse} apply FC layer along  horizontal, vertical, and channel directions, respectively, 
in order to reduce the number of parameters and computation cost. 
However, the parameters of FC layer are still determined by the input resolution, so it is hard to handle different image scales. 
CycleMLP \cite{cyclemlp} addresses such problem with padding,
but it only focuses on global information, 
ignoring local inductive bias. 
Meanwhile, the ability of MLP-Like architecture for video modeling has not been explored. 

On the contrary, our MorphMLP can cope with diverse scales via splitting the sequence of tokens into  chunks. 
Furthermore, it is able to effectively capture local to global information by gradually expanding chunk length. More importantly, we are the first to build MLP-Like architecture on videos to explore its generalization ability as a new paradigm of versatile backbone.









\section{Method}

In this section, we present our MorphMLP.
We first introduce the two critical components of MorphMLP, $\tt{MorphFC_s}$ and $\tt{MorphFC_t}$. Then, we illustrate how to build efficient spatial-temporal  MorphMLP block. 
Finally, the overall spatial-temporal network architecture and its adaption to image domain are provided.

\subsection{MorphFC for Spatial Modeling}

\begin{figure}[t]
\centering
\begin{minipage}[t]{0.52\linewidth}
\centering
\includegraphics[width=\textwidth]{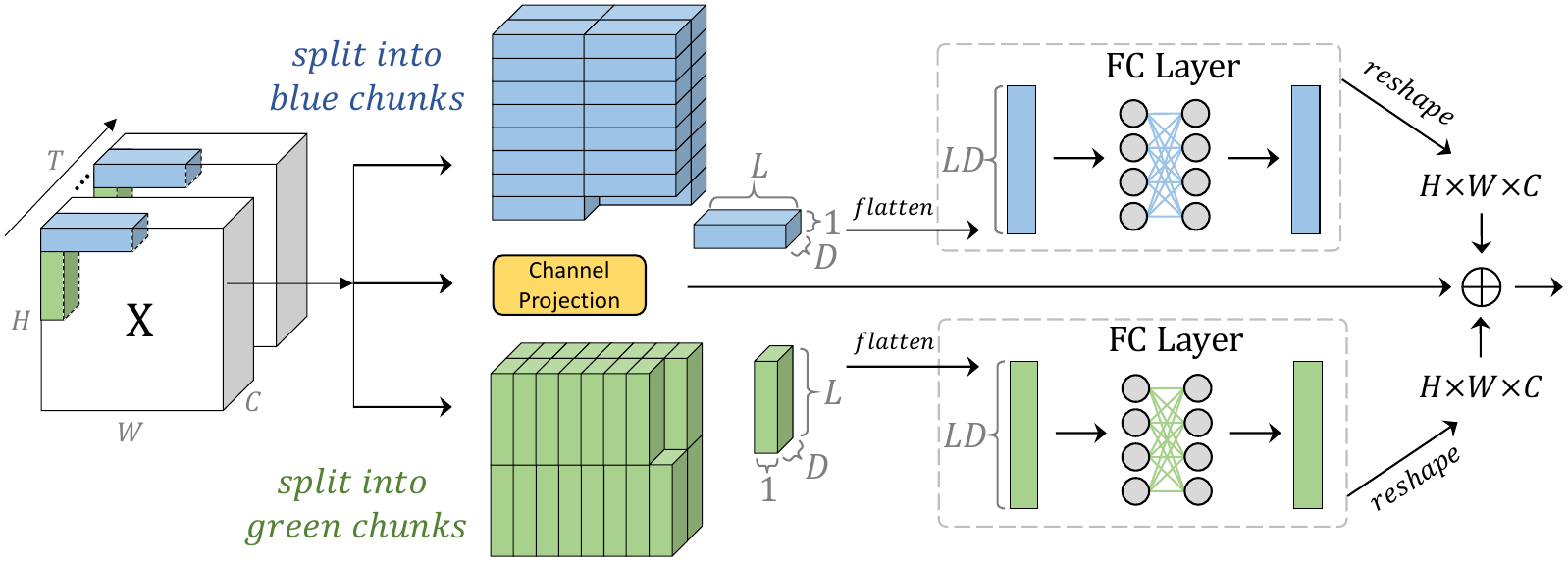}
\caption{$\tt MorphFC_s$ on the spatial dimension. Note that chunk length L hierarchically expands as network goes deeper.}
\label{fig:spatial}
\end{minipage}
\hspace{0cm}
\begin{minipage}[t]{0.45\linewidth}
\centering
\includegraphics[width=\textwidth]{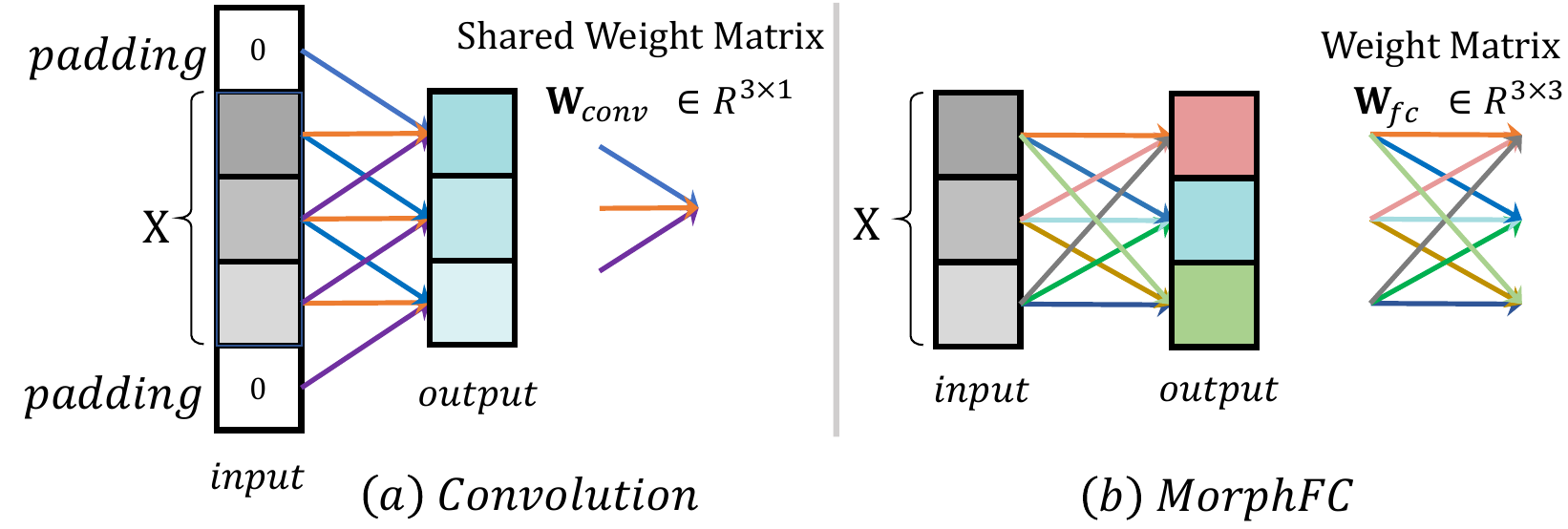}
\caption{Comparison with the typical convolution.}
\label{fig:vsconv}
\end{minipage}
\end{figure}

As discussed above, mining core semantics is critical to video recognition. Typical CNN and previous MLP-Like architectures only focus on either local or global information modeling thus they fail to do that. To tackle this challenge, we propose a novel $\tt MorphFC_s$ layer that can hierarchical expand the receptive field of FC  and make it operate from small to big regions.
Our $\tt MorphFC_s$ processes each frame of video independently in horizontal and vertical pathways. We take the horizontal one (blue chunks in Fig.  \ref{fig:spatial}) for example.

Specifically, given one frame of input videos $\mathbf{X}\in \mathbb{R}^{HW \times C}$ that has been projected into a sequence of tokens, we first split $\mathbf{X}$ along horizontal direction. We set chunk length to $L$ and thus obtain  $\mathbf{X}_{i}\in \mathbb{R}^{L\times C}$, where $ i \in \{1,...,HW/L\} $. Furthermore, to reduce computation cost, we also split each  $\mathbf{X}_{i}$ into multiple groups along channel dimension, where each group has $D$ channels.
Thus we get split chunks,
and each single chunk is  $\mathbf{X}_{i}^{k}\in \mathbb{R}^{LD}$, where $ k \in \{1,...,C/D\}$.
Next, we flatten each chunk into 1D vector and apply a  FC weight matrix $\mathbf{W} \in \mathbb{R}^{LD\times LD} $ to transform each chunk, yielding
\begin{equation}
\mathbf{Y}_{i}^{k}= \mathbf{X}_{i}^{k}\mathbf{W}.
    \end{equation}
After feature transformation, we reshape all chunks  $\mathbf{Y}_{i}^{k}$ back to the original dimension $\mathbf{Y} \in \mathbb{R}^{H\times W \times C}$. The vertical way (green chunks in Fig.  \ref{fig:spatial}) does likewise except splitting the sequence of tokens along vertical direction.  To make communication among groups along channel dimension, we also apply a FC layer to process each token individually.   Finally, we get the output by element-wise summing horizontal, vertical, and channel features together. The chunks length $L$ hierarchically increases as the network deepens, thereby enabling the FC filter to discover more core semantics progressively from small to big spatial region.


\noindent 
\textbf{Difference between our $MorphFC_s$ and convolution.}  \textbf{(i)} Typical convolution utilizes fixed small kernel size (e.g., 3$\times$3), which only aggregates local context. On the contrary, the chunks lengths in $\tt MorphFC_s$  hierarchically  increase as the network deepens, which can model short-to-long range information progressively.
\textbf{(ii)} Convolution uses sliding windows to obtain overlapping tokens, which requires
cumbersome operations, including unfold, reshape and fold.  
In contrast, we simply reshape the feature map to obtain our chunks with non-overlapping tokens. 
\textbf{(iii)} As shown in Fig.  \ref{fig:vsconv}, given a 1$\times$3 input, to get the 1$\times$3 output,  the convolution kernel of 1$\times$3 window size needs to slide three times, and each 1$\times$1 output is generated by the shared weight matrix $\mathbf{W}_{conv} \in \mathbb{R}^{3\times 1 }$. In contrast, FC layer applies weight matrix $\mathbf{W}_{fc} \in \mathbb{R}^{3\times 3 }$ to the input yielding 1$\times$3 ouput. Each 1$\times$1 output  is equivalent to being generated by non-shared weight matrix $\mathbf{W} \in \mathbb{R}^{3\times 1 }$, which brings more flexible spatial encoding than convolution.  

\begin{figure*}[t]
    \begin{center}
        \includegraphics[width=.95\textwidth]{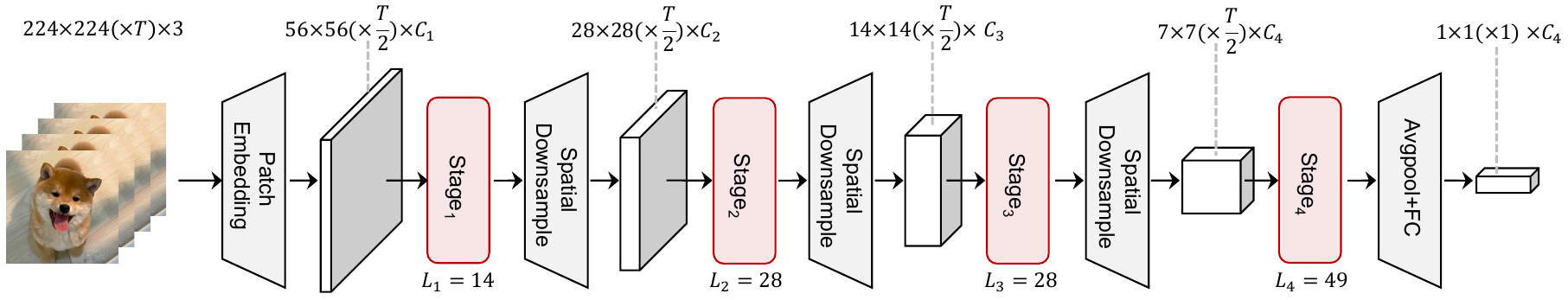}
        \end{center}
     
        \caption{
        Architecture of our MorphMLP $L$ means chuck length.}

    \label{fig:framework}
\end{figure*}

\noindent\textbf{Comparisons with ViP \cite{vip}.} Our design is related to the well-known ViP designed for image domain,
which also leverages the multi-branch features in spatial modeling. Hence, we further discuss the differences. 
\textbf{(i)} The
FC filters of whole ViP network have the fixed size and receptive field, thus they only capture global information. 
On the contrary, our FC filters are morphable, as shown in Fig. \ref{fig:res}. In shallow layers, they
have small size to model local structure, while in deeper layers, they gradually
change to large size to model long-range information. Hence, ours can discover more detailed semantics by progressively operating FC from small to big spatial region. 
\textbf{(ii)}As shown in Fig. \ref{fig:framework}, at the network level, ours have hierarchical downsampling after each stage but ViP does not. 
\textbf{(iii)} As ViP paper said, ViP is hard to transfer to downstream tasks
i.e. segmentation with spatial resolution 2048$\times$512,
since its filter size is always equal to the
height/weight of features. 
But it is easy for ours,
because the filter size is equal to pre-defined chunk size in the pre-training.




\begin{figure}[t]
\begin{center}
\includegraphics[width=0.6\textwidth]{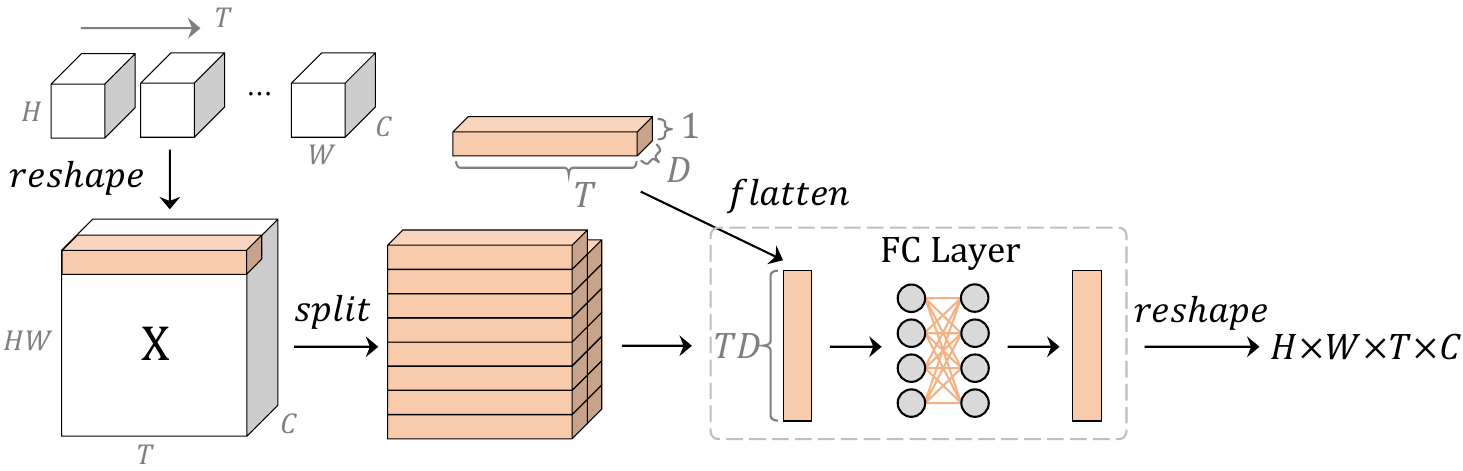}
\end{center}

\caption{
$\tt MorphFC_{t}$ on the temporal dimension.
}

\label{fig:t_cmblock}
\end{figure}
\subsection{MorphFC$_t$ on Temporal Modeling}
 In addition to the horizontal and vertical pathways in $\tt MorphFC_s$, we introduce another temporal pathway $\tt MorphFC_{t}$. It aims at capturing long-term temporal information using the simple FC layer with low computation cost. 
 Specifically, as shown in Fig.  \ref{fig:t_cmblock}, given an input video clip tokens  $\mathbf{X}\in \mathbb{R}^{H\times W\times T \times C}$, we first split X into a couple of groups  along channel dimension ($D$ channels in each group) to reduce computation cost and get  $\mathbf{X}^{k}\in \mathbb{R}^{H\times W\times T \times D}$, where $ k \in \{1,...,C/D\}$. For each spatial position s, we concatenate features across all frames into a chunk  $\mathbf{X}^{k}_{s}\in \mathbb{R}^{ TD}$, where $ s \in \{1,...,HW\}$. Then we apply a $FC$ matrix $\mathbf{W} \in \mathbb{R}^{ TD \times TD}$,  to transform temporal features and get 
\begin{equation}
\mathbf{Y}^{k}_{s} = \mathbf{X}^{k}_{s}\mathbf{W}.
    \end{equation}
Finally, we reshape all chunks $\mathbf{Y}^{k}_{s} \in \mathbb{R}^{TD}$ back to original tokens  dimension and output $\mathbf{Y}\in \mathbb{R}^{H\times W\times T \times C}$. In this way, the FC filter can  simply aggregate token relations along time dimension in the chunk to model temporal dependencies.

\begin{figure*}[t]
    \begin{center}
        \includegraphics[width=.98\textwidth]{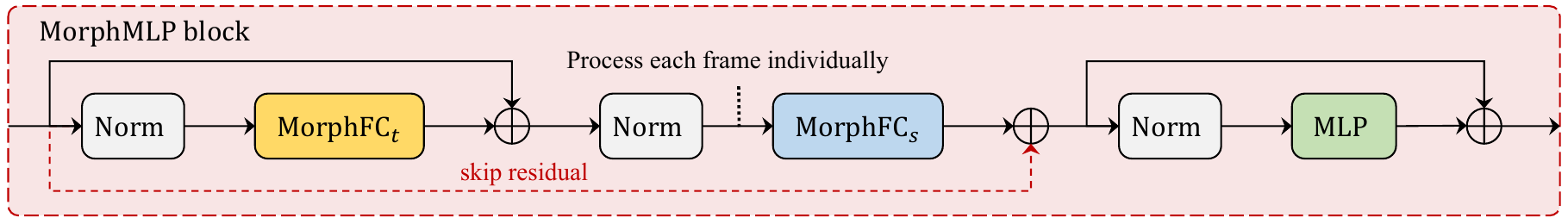}
        \end{center}
      
        \caption{
        Spatial-Temporal MorphMLP  Block.
        }
 
    \label{fig:video_block}
\end{figure*}

\noindent\textbf{Spatial-Temporal MorphMLP block.}
Based on the  $\tt MorphFC_s$ and  $\tt MorphFC_{t}$, we propose a factorized spatial-temporal  MorphMLP block in the video domain for efficient video representation learning. As shown in Fig. \ref{fig:video_block}, our MorphMLP block contains $\tt MorphFC_{t}$, $\tt MorphFC_s$ and MLP \cite{transformer} modules in a sequential order.  
On one hand,
it is difficult for joint spatial-temporal optimization \cite{timesformer}.
On the other hand, factorizing spatial and temporal modeling is able to reduce the computation cost significantly. Therefore,
we place temporal and spatial $\tt MorphFC_s$ layers in the sequential style.
The LN \cite{ln} layer is applied before each module, and the standard residual connections are used after $\tt MorphFC_{t}$ and MLP module.
Instead of applying a standard residual connection \cite{resnet} after $\tt MorphFC_s$,
we add a skip residual connection (red line) between the original input and output features from $\tt MorphFC_s$ layer. We found that such a connection can make training more stable.

\noindent
\subsection{Network architecture}
For video recognition, as shown in Fig. \ref{fig:framework}, we hierarchically stack spatial-temporal MorphMLP blocks  to build up our network. Given an video sequence $\mathbf{X}\in \mathbb{R}^{ H\times W \times T\times 3}$, taking $H$=$W$=224 for example, our MorphMLP backbone first performs patch embedding  on the video clip and gets a sequence of tokens with dimension  56$\times$56$\times T/2 \times C_1$. Then, we have four sequential stages and each of them contains a couple of MorphMLP blocks. The feature size remains unchanged as passing through layers inside the same stage. At the end of each stage excluding the last one, we expand the channel dimension and downsample the spatial resolution of features by ratio 2.

Note that we set chunk lengths of $\tt MorphFC_s$ to be 14, 28, 28, 49 for stage 1-4, respectively. Horizontal/vertical chunks  with lengths 14, 28, 28, 49 of stage 1-4 can cover quarter, one, two, all rows/columns of feature maps of stage 1-4, respectively. In shallow layers,  our network can learn detailed representation from the local spatial context in small chunk length, e.g., length 14 for 56$\times$56$\times C_1$
feature map. In deep layers, our network can capture long-range information from the global semantic context in considerable chunk length, e.g., length 49 for  7$\times $7$\times C_4$ feature map. With downsampling the spatial resolution and expanding chunk length as the network goes deeper, our MorphMLP is capable of discovering more core semantics progressively by operating the FC filter from small to big spatial regions.

We provide two model variants for the video recognition depending on the number of MorphMLP blocks in four stages: $\{3, 4, 9, 3\}$ for MorphMLP-Small(S) and  $\{4, 6, 15, 4\}$  for MorphMLP-Base(B).
The numbers of channels of four stages are  $\{112,224,392,784\}$.

\noindent{\textbf{Adaption to Imgae domain.}} Additionally,
for image-domain architecture,  we simply exclude the temporal dimension and drop $\tt MorphFC_t$ in the MorphMLP block. 
In addition to small and base settings, 
we provide two extra model variants for image domain,
depending on the number of MorphMLP blocks in four stages,
i.e.,
$\{3, 4, 7, 3\}$ for MorphMLP-Tiny(T) 
and  
$\{4, 8, 18, 6\}$  for MorphMLP-Large(L).

\section{Experiment}
In this section, we first examine the performance of MorphMLP  and evaluate its spatiotemporal effectiveness on Kinetics-400\cite{k400}, and Something-Something V1\&V2 \cite{sth} datasets. For fair comparisons and due to GPU resources limitation, we only report MorphMLP-S and B for video classification.
Then we verify the effectiveness of its adaption to image domain, including ImageNet-1K \cite{imagenet} image classification and ADE20K\cite{ade20k} semantic segmentation.

\begin{table}[t]
\centering
\caption{Comparisons with the state-of-the-art on Kinetics-400\cite{k400}. Our MorphMLP  achieves outstanding results with much fewer computation costs. For example, compared with VideoSwin-T, our MorphMLP-S only requires 2$\times$ fewer GFLOPs but gets 0.9\% accuracy improvement (79.7\% \textit{vs.} 78.8\%).}

\centering
    \setlength{\tabcolsep}{8.0pt}
    \resizebox{0.80\columnwidth}{!}{
	\begin{tabular}{l|l|l|l|cc}
	    \Xhline{1.0pt}
		\multirow{2}*{Method} & \multirow{2}*{Pretrain} & \multirow{2}*{\#Frame} & \multirow{2}*{GFLOPs} & \multicolumn{2}{c}{K400}  \\
		 ~  & ~ & ~ & &~ Top-1 & Top-5 \\
	    \Xhline{0.7pt}
		\multicolumn{6}{c}{Self-Attention Free -- CNN} \\
	    \Xhline{0.7pt}
		SlowFast R101\cite{slowfast}  & - & (16+64)$\times$3$\times$10 & 6390 & 78.9  & 93.5  \\
		CorrNet-101\cite{corrnet}  & - & 32$\times$ 3$\times$10  & 6720& 79.2  &- \\
		ip-CSN\cite{csn}  & Sports1M & 32$\times$ 3$\times$10  & 3264& 79.2  & 93.8 \\
		X3D-XL\cite{x3d}    &- & 16$\times$3$\times$10 & 1452& 79.1    &93.9 \\
		SmallBig$_{EN}$\cite{smallbig}  & IN-1K & (8+32)$\times$3$\times$4 &  5700& 78.7    &93.7 \\
		TDN$_{EN}$\cite{tdn}   & IN-1K & (8+16)$\times$3$\times$10 &  5940& 79.4    &94.4 \\
		CT-Net$_{EN}$\cite{ct_net}  & IN-1K & (16+16)$\times$3$\times$4 &  2641& 79.8 &94.2 \\
	    \Xhline{0.7pt}
    	\multicolumn{6}{c}{Self-Attention Based -- Transformer} \\
	    \Xhline{0.7pt}
		Timesformer-L\cite{timesformer} &IN-21K & 96$\times$3$\times$1 & 7140& 80.7 &94.7\\
		VidTr-L\cite{vidtr}  &IN-21K & 32$\times$3$\times$10 & 11760  & 79.1  &93.9  \\
		ViViT-L\cite{vivit}  &IN-21K & 16$\times$3$\times$4 & 17357 & 80.6 &94.7  \\
		X-ViT\cite{x_vit} &IN-21K & 16$\times$3$\times$1 & 850 & 80.2  &94.7\\
		Mformer\cite{motionformer} &IN-21K & 32$\times$3$\times$10 & 11085  &80.2   &94.8  \\
		Mformer-L\cite{motionformer} &IN-21K & 32$\times$3$\times$10 & 35550  &80.2   &94.8  \\
		MViT-B,16$\times$4 \cite{mvit}  &- & 16$\times$1$\times$5 & 355 & 78.4  &93.5 \\
		MViT-B,32$\times$3 \cite{mvit}  &- & 32$\times$1$\times$5 & 850 & 80.2  &94.4  \\
		VideoSwin-T\cite{video_swin}  &IN-1K & 32$\times$3$\times$4 & 1056 & 78.8  &93.6  \\
		VideoSwin-B\cite{video_swin}  &IN-1K & 32$\times$3$\times$4 & 3384 & 80.6  &94.6  \\
	    \Xhline{0.7pt}
        \multicolumn{6}{c}{Self-Attention Free -- MLP-Like} \\
	    \Xhline{0.7pt}
		MorphMLP-S & IN-1K & 16$\times$1$\times$4 & 268 &78.7 &93.8  \\
		MorphMLP-S & IN-1K & 32$\times$1$\times$4 &532 &79.7 &94.2
		\\
		MorphMLP-B & IN-1K & 16$\times$1$\times$4 &392 &79.5 &94.4
		\\
		MorphMLP-B & IN-1K & 32$\times$1$\times$4 &788 &\textbf{80.8} &\textbf{94.9}
		\\
	\Xhline{1.0pt}
	\end{tabular}
    }
    \label{sota_k400}

\end{table}

\subsection{Video Classification on Kinetics-400}

\noindent
\textbf{Settings.} Kinetics-400\cite{k400} is a large-scale scene-related video benchmark.
It contains around 240K training videos and about 20K validation videos in 400 classes.
Our code heavily relies on \texttt{PySlowFast} \cite{pyslowfast} repository and the training recipe mainly follows MViT\cite{mvit}.
We directly load the parameters of $\tt MorphFC_s$ pre-trained on ImageNet and randomly initialize the parameters of $\tt MorphFC_{t}$ in the video domain.
We adopt a dense sampling strategy \cite{non_local} and AdamW optimizer to train the whole network.
The warm-up epoch, total epoch, batch size, base learning rate, and weight decay are 10, 60, 64, 2e-4, and 0.05 respectively.
We utilize the stochastic depth rates 0.1 and 0.3 for MorphMLP-S and B.

\noindent
\textbf{Results.} 
As shown in Table \ref{sota_k400}, our method achieves  outstanding performance with fewer computation costs.  Compared with CNN models such as SlowFast\cite{slowfast}, our MorphMLP requires 8$\times$ fewer GFLOPS but  achieves 1.9\% accuracy improvement (80.8\% \textit{vs.} 78.9\%). With only ImageNet-1K pre-training, our method surpasses most of the  self-attention based Transformer backbones with larger dataset pre-training.
For example, compared with ViViT-L\cite{vivit} pre-trained on ImageNet-21K, our MorphMLP obtains better
performance with 20$\times$ fewer computations. When our model is scaled larger, the accuracy increases as well. Since the computation cost is relatively low, our method still has great potential for better performance. It demonstrates that our MorphMLP is a strong MLP-Like backbone for video recognition.

\begin{table*}[t]
\centering
    \begin{minipage}[t]{1\linewidth}
    \caption{Comparisons with the SOTA on SSV2 \cite{sth}. Our MorphMLP outperforms previous sota Transformers and CNNs with IN-1K pretraining only.
    }
\centering
    \setlength{\tabcolsep}{7.0pt}
    \resizebox{0.80\textwidth}{!}{
	\begin{tabular}{l|l|l|l|cc}
        \Xhline{1.0pt}
		\multirow{2}*{Method} & \multirow{2}*{Pretrain} & \multirow{2}*{\#Frame} & \multirow{2}*{GFLOPs} & \multicolumn{2}{c}{SSV2}  \\
		 ~ & ~ & ~  & ~ & Top-1 & Top-5 \\
        \Xhline{0.7pt}
        \multicolumn{6}{c}{Self-Attention Free -- CNN} \\
        \Xhline{0.7pt}
		SlowFast R50\cite{slowfast}  &K400 & (8+32)$\times$3$\times$1  & 197 & 61.7  &46.6    \\
		TSM\cite{tsm}  &K400 & 16$\times$3$\times$2 & 374 & 63.4  &88.5  \\
		STM\cite{stm}  &IN-1K & 16$\times$3$\times$10        & 1995 & 64.2  &89.8 \\
		bLVNet\cite{more_is_less}  &IN-1K & 32$\times$3$\times$10 & 3870& 65.2  &90.3\\
		TEA\cite{tea}   &IN-1K & 16$\times$3$\times$10  & 2100 & 65.1  &-\\
		CT-Net\cite{ct_net}   &IN-1K & 16$\times$3$\times$2  & 450 & 65.9
		&90.1\\
        \Xhline{0.7pt}
        \multicolumn{6}{c}{Self-Attention Based -- Transformer} \\
        \Xhline{0.7pt}
		Timesformer\cite{timesformer}     &IN-21K & 16$\times$3$\times$1 & 5109  & 62.5 &-\\
		VidTr-L\cite{vidtr}  &IN-21K+K400 & 32$\times$3$\times$10 & 10530  & 60.2  &-  \\
		ViViT-L\cite{vivit}  &IN-21K+K400 & 16$\times$3$\times$4 & 11892 & 65.4  &89.8  \\
		X-ViT\cite{x_vit} &IN-21K & 32$\times$3$\times$1 & 1269 & 65.4  &90.7\\
		Mformer\cite{motionformer}  &IN-21K+K400 & 16$\times$3$\times$1 & 1110 &66.5   &90.1  \\
		Mformer-L\cite{motionformer}  &IN-21K+K400 & 32$\times$3$\times$1 & 3555& 68.1  &91.2  \\
		MViT-B,16$\times$4\cite{mvit}  &K400 & 16$\times$3$\times$1 & 510 & 67.1  &90.8 \\
		MViT-B,32$\times$3\cite{mvit}  &K400 & 32$\times$3$\times$1 & 1365  & 67.7  &90.9  \\
		MViT-B-24,32$\times$3\cite{mvit}  &K600 & 32$\times$3$\times$1 & 708  & 68.7  &91.5  \\
        \Xhline{0.7pt}
    	\multicolumn{6}{c}{Self-Attention Free -- MLP-like} \\
        \Xhline{0.7pt}
		MorphMLP-S & IN-1K & 16$\times$3$\times$1 & 201  &67.1 & 90.9  \\
		MorphMLP-S & IN-1K & 32$\times$3$\times$1 & 405 &68.3 &91.3
		\\
		MorphMLP-B & IN-1K & 16$\times$3$\times$1 & 294  &67.6 &91.3
		\\
		MorphMLP-B& IN-1K & 32$\times$3$\times$1 & 591  &\textbf{70.1} &\textbf{92.8}
		\\
        \Xhline{1.0pt}
	\end{tabular}}
    \label{sota_sthv2}
    \end{minipage}

    \begin{minipage}[t]{1\linewidth}
  \caption{Comparisons with the state-of-the-art on Something-Something V1 \cite{sth}.}
    \setlength{\tabcolsep}{10.0pt}
    \centering
    \resizebox{0.8\columnwidth}{!}{
	\begin{tabular}{l|l|l|l|cc}
        \Xhline{1.0pt}
		\multirow{2}*{Method} & \multirow{2}*{Pretrain} & \multirow{2}*{\#Frame} & \multirow{2}*{GFLOPs} & \multicolumn{2}{c}{SSV1}  \\
		 ~ & ~ & ~  & ~ & Top-1 & Top-5 \\
        \Xhline{0.7pt}
		I3D\cite{video_as}  &IN-1K+K400 & 32$\times$3$\times$2 & 918 & 41.6  &72.2  \\
		NLI3D\cite{video_as}   &IN-1K+K400 & 32$\times$3$\times$2 & 1008 & 44.4  &76.0 \\
		NLI3D+GCN\cite{video_as}   &IN-1K+K400 & 32$\times$3$\times$2 & 1818 &46.1   &76.8 \\
		TSM\cite{tsm}   &IN-1K+K400 & 16$\times$1$\times$1 & 65   & 47.2  &77.1\\
		SmallBig\cite{smallbig}   &IN-1K & 16$\times$1$\times$1  &105   &49.3   &79.5 \\
		TEINet\cite{tei}   &IN-1K & 16$\times$3$\times$10 & 1980 & 51.0  &-  \\
		TEA\cite{tea}   &IN-1K & 16$\times$3$\times$10 & 2100 & 52.3  & 81.9  \\
		CT-NET\cite{ct_net}   &IN-1K & 16$\times$3$\times$2 & 447 & 53.4  & 81.7  \\
        \Xhline{0.7pt}
		MorphMLP-S & IN-1K  & 16$\times$1$\times$1 & 67  & 50.6 &78.0  \\
		MorphMLP-S & IN-1K  & 16$\times$3$\times$1 & 201  & 53.9 & 81.3  \\
		MorphMLP-B & IN-1K  & 16$\times$3$\times$1 & 294  & 55.5 &82.4  \\
		MorphMLP-B & IN-1K  & 32$\times$3$\times$1 & 591  & \textbf{57.4} &\textbf{84.5}  \\
    \Xhline{1.0pt}
	\end{tabular}
	 
    }
   
    \label{sota_sthv1}

    \end{minipage}

\end{table*}

\subsection{Video Classification on Something-Something}

\noindent \textbf{Settings.} Something-Something \cite{sth} is another large-scale dataset,
in which the temporal relationship modeling is critical for action understanding.
It includes two versions, 
i.e., V1 and V2, 
both of which contain plentiful videos over 174 categories.
We adopt the same training setting as used for Kinetics-400,
except that a random horizontal flip is not applied.
We utilize the sparse sampling strategy.
The warm-up epoch, total epoch, batch size, base learning rate, and weight decay are 5, 50, 64, 4e-4, and 0.05, respectively.
We set the stochastic depth rates to be 0.3 and 0.6 for Morph-S and B respectively.

\noindent
\textbf{Results.} The comparison results on Something V2\&V1 are  shown in Table \ref{sota_sthv2} and Table \ref{sota_sthv1}  respectively. For SSV2, CNN  architectures perform worse than Transformer architectures since they are limited to capturing local spatial and temporal information and struggle to model long-term dependencies. Transformer architectures can achieve better results, but they heavily rely on large-scale dataset pre-training which requires high computation. Compared with CT-Net\cite{ct_net}, our MorphMLP can reduce 2.5$\times$ computation but achieves 1.2\% accuracy gain.   Compared with the-state-of-art method MViT\cite{mvit}, which is pre-trained on large video dataset Kinetics-600, our MorphMLP only pre-trained on  ImageNet-1K can obtain better performance (70.1\% \textit{vs.}  68.7\%) with smaller GFLOPS (591G \textit{vs.}  708G).  For SSV1, our MorphMLP also achieves outstanding results. 

The superior results of our method on this dataset can be attributed to our unique progressively  core semantics discovering manner and efficient spatial-temporal block design in MorpMLP. Table \ref{ablation_length} and \ref{ablation_residual} can also demonstrate our point. Note that even if we do not add any complicated and unique temporal attention operation, our simple method can achieve such great performance. This indicates that our model can serve as a strong backbone for further improvement.

\subsection{Image Classification on ImageNet-1K}

\begin{table}[t]
    \centering
        \caption{ImageNet-1K results. As shown in (a), our method achieves the best performance among SOTA MLP-Like architectures. From (b), we can see that our MorphMLP also achieves the comparable results with SOTA self-attention based and hybrid models even with small computation. }
      
    \begin{subtable}[t]{0.5\linewidth}
           \caption{Comparisons with MLP-Like models.}

        \centering
        \setlength{\tabcolsep}{5.5pt}
        \resizebox{0.94\textwidth}{!}{
       \begin{tabular}{l | c c|c}
\Xhline{1.0pt}
Model & Param & FLOPs   &  Top-1 \\
\hline

Mixer-B/16\cite{mixer}          & 59M  & 12.7G & 76.4  \\
Mixer-B/16$^\dagger$\cite{mixer}& 59M  & 12.7G & 77.3  \\ \hline

ResMLP-S12\cite{resmlp}              & 15M  & 3.0G  & 76.6 \\
ResMLP-S24\cite{resmlp}              & 30M  & 6.0G  & 79.4 \\
ResMLP-B24 \cite{resmlp}             & 116M & 23.0G & 81.0 \\ \hline

gMLP-Ti\cite{gmlp}                   & 6M   & 1.4G  & 72.3 \\
gMLP-S \cite{gmlp}                   & 20M  & 4.5G  & 79.6 \\
gMLP-B  \cite{gmlp}                  & 73M  & 15.8G & 81.6 \\ \hline

S$^2$-MLP-wide \cite{smlp}        & 71M  & 14.0G & 80.0 \\
S$^2$-MLP-deep \cite{smlp}        & 51M  & 10.5G & 80.7 \\ \hline

ViP-Small/7 \cite{vip}    & 25M   & 6.9G  & 81.5 \\
ViP-Medium/7\cite{vip}     & 55M   & 16.3G & 82.7 \\
ViP-Large/7 \cite{vip}     & 88M   & 24.4G & \text{83.2} \\ \hline

AS-MLP-T \cite{asmlp}          & 28M   & 4.4G  & 81.3 \\
AS-MLP-S \cite{asmlp}          & 50M   & 8.5G  & 83.1 \\
AS-MLP-B \cite{asmlp}          & 88M   & 15.2G & 83.3 

\\ \hline

CycleMLP-B2\cite{cyclemlp}                  & 27M   & 3.9G  & 81.6 \\
CycleMLP--B3\cite{cyclemlp}                    & 38M   & 6.9G  & 82.6 \\
CycleMLP--B4\cite{cyclemlp}                        & 52M   & 10.1G & 83.0 \\
CycleMLP--B5\cite{cyclemlp}                        & 76M   & 12.3G & 83.1 \\ 
\hline
	\rowcolor{gray!20}
  MorphMLP-T   & 23M & 3.9G  & 81.6 \\ 
	\rowcolor{gray!20}
  MorphMLP-S  & 38M &6.9G & 82.6 \\ 
	\rowcolor{gray!20}
MorphMLP-B  & 58M & 10.2G & 83.2  \\ 
	\rowcolor{gray!20}
MorphMLP-L  &76M &12.5G &\textbf{83.4}\\
\Xhline{1.0pt}
\end{tabular}
        }
    \label{mlp_in1k}
    \end{subtable}
    \begin{subtable}[t]{0.48\linewidth}
       \caption{Comparisons with SOTA models.}

        \centering
        \setlength{\tabcolsep}{1pt}
        \resizebox{1.0\textwidth}{!}{
           \begin{tabular}{l | c |c c c|c}
\Xhline{1.0pt}
Model & Family & Scale & Param & FLOPs   &  Top-1 \\
\hline

ResNet50 \cite{resnet}                  &  CNN  & 224$^2$ & 26M & 4.1G  & 79.2 \\
DeiT-S \cite{deit}           & Trans & 224$^2$ & 22M & 4.6G  & 79.8 \\
ResNest50\cite{resnest}  &CNN&224  &28M &4.3G &80.6\\
T2T-ViT-14 \cite{t2t} & Trans & 224$^2$ & 22M & 4.8G  & 81.5 \\
PVT-S \cite{pvt}                & Trans & 224$^2$ & 25M & 3.8G  & 79.8 \\
Swin-T  \cite{swin}                 & Trans & 224$^2$ & 29M & 4.5G  & 81.3 \\
GFNet-H-S \cite{gfnet}              & FFT   & 224$^2$ & 32M & 4.5G  & 81.5 \\
BoT-S1-50 \cite{botnet}      & Hybrid& 224$^2$ & 21M & 4.3G  & 79.1 \\
CoAtNet-0\cite{coatnet}     &Hybrid &224$^2$ &23M&4.2G &81.6 \\
	\rowcolor{gray!20}
MorphMLP-T        & MLP   & 224$^2$ & 23M & 3.9G  & \textbf{81.6} \\ \hline

ResNet101 \cite{resnet}                &  CNN  & 224$^2$ & 45M & 7.9G & 79.8 \\
ResNest101\cite{resnest}& CNN & 224$^2$ & 48M &8.0G &82.0\\
RegNetY-8G \cite{regnet}  &  CNN  & 224$^2$ & 39M & 8.0G & 81.7 \\
T2T-ViT-19 \cite{t2t} & Tran & 224$^2$ & 39M & 8.5G & 81.9 \\
PVT-M \cite{pvt}                & Trans & 224$^2$ & 44M & 6.7G & 81.2 \\
BoT-S1-59 \cite{botnet}     & Hybrid& 224$^2$ & 34M & 7.3G & 81.7  \\
CoAtNet-1\cite{coatnet}     &Hybrid &224$^2$&42M&8.4G &\textbf{83.3} \\
	\rowcolor{gray!20}
MorphMLP-S                             & MLP   & 224$^2$ & 38M & 6.9G & 82.6 \\ \hline

ViT-B/16 \cite{deit}       & Trans & 384$^2$ & 86M & 55.4G & 77.9 \\
DeiT-B \cite{deit}         & Trans & 224$^2$ & 86M & 17.5G & 81.8 \\
DeiT-B \cite{deit}         & Trans & 384$^2$ & 86M & 55.4G & 83.1 \\
T2T-ViT-24 \cite{t2t} & Tran & 224$^2$ & 64M & 13.8G & 82.3 \\
Swin-B \cite{swin}         & Trans & 224$^2$ & 88M & 15.4G & 83.4 \\
CaiT-S36  \cite{cait}        & Trans & 224$^2$ & 68M & 13.9G & 83.3   \\
	\rowcolor{gray!20}
MorphMLP-L  & MLP   & 224$^2$ & 76M & 12.5G & \textbf{83.4} \\ 
\Xhline{1.0pt}
     \end{tabular}  }
      \label{sota_in1k}
    \end{subtable}
    \label{in1k}
    
\end{table}

\noindent
\textbf{Settings.} We train our models from scratch on the ImageNet-1K dataset \cite{imagenet},
which consists of 1.2M training images and 50K validation images from 1,000 categories.
Our code is implemented based on \texttt{DeiT}\cite{deit} repository, and we follow the same training strategy proposed in DeiT\cite{deit},
including strong data augmentation and regularization. 
Stochastic depth rates are set to be 0.1, 0.1, 0.2, 0.3 for our 4 model variants.
We adopt AdamW \cite{adamw} optimizer with cosine learning rate schedule \cite{cosine} for 300 epochs,
while the first 20 epochs are used for linear warm-up\cite{warmup}. 
The total batch size, weight decay, and initial learning rate are set to 1024, 5$\times$10$^{-2}$ and 0.01 respectively.

\noindent
\textbf{Results.} As shown in Table \ref{mlp_in1k}, our MorphMLP outperforms the state-of-the-art MLP-Like architectures. Compared with ViP-S\cite{vip}, our method can get much higher accuracy (82.6\% \textit{vs.}  81.5\%) with similar GFLOPS (7.0G \textit{vs.} 6.9G). This demonstrates the effectiveness of our progressively short-to-long range  pattern. In Table \ref{sota_in1k}, our MorphMLP can achieve competitive results with popular self-attention based models. Compared with other tiny models, e.g., Swin-T\cite{swin}, our method can achieve better results (81.6\% \textit{vs.}  81.3\%) with fewer parameters and GFLOPS (23M \textit{vs.}  29M, 3.9G \textit{vs.}  4.5G). As for larger settings, our method can achieve comparable result to Swin-B \cite{swin} with fewer GFLOPS.

\subsection{Semantic Segmentation on ADE20K}

\begin{table}[t]
    \centering
\caption{Semantic segmentation with Semantic FPN \cite{fpn} on ADE20K \cite{ade20k} val.}
    \setlength{\tabcolsep}{12pt}
    \centering
	\resizebox{0.7\columnwidth}{!}{
    \begin{tabular}{l|l|c|c}
	\Xhline{1.0pt}
    Method & Arch & \#Param.(M)    &  mIoU \\
	\Xhline{0.7pt}
    ResNet50\cite{resnet} & CNN & 28.5 & 36.7 \\
    PVT-S\cite{pvt} & Trans & 28.2 & 39.8 \\
    Swin-T\cite{swin} & Trans & 31.9 & 41.5 \\
	GFNet-H-T\cite{gfnet} & FFN & 26.6 & 41.0 \\
	CycleMLP-B2 \cite{cyclemlp}             &MLP-Like        & 30.6 & 42.4 \\
	\rowcolor{gray!20}
	MorphMLP-T & MLP-Like & 26.4 &\textbf{43.0} \\
	\Xhline{0.7pt}
    ResNet101\cite{resnet} & CNN & 47.5 & 38.8\\
    ResNeXt101-32$\times$4d\cite{resnext} & CNN & 47.1 & 39.7 \\
    PVT-M\cite{pvt} & Trans & 48.0 & 41.6 \\
    GFNet-H-S\cite{gfnet} & FFN & 47.5 & 42.5 \\
    CycleMLP-B3\cite{cyclemlp}     &MLP-Like                   & 42.1 & 44.5 \\
	\rowcolor{gray!20}
     MorphMLP-S & MLP-Like & 41.0 & \textbf{44.7} \\
	\Xhline{0.7pt}
    PVT-L\cite{pvt}& Trans & 65.1 & 42.1 \\
    Swin-S\cite{swin}& Trans & 53.2 & 45.2 \\
    CycleMLP-B4 \cite{cyclemlp}      &MLP-Like              & 55.6 & 45.1 \\
	\rowcolor{gray!20}
    MorphMLP-B & MLP-Like& 59.3 &\textbf{45.9} \\
	\Xhline{1.0pt}
    \end{tabular}}
   
	\label{sota_seg}
\end{table}

\noindent
\textbf{Settings.} We conduct semantic segmentation experiments on ADE20K\cite{ade20k},
which consists of 20K training images and 2K validation images over 150 semantic categories.
Our code is based on \texttt{mmsegmentation} \cite{mmseg} and we follow the experiment setting used in PVT\cite{pvt}.
We simply apply Semantic FPN \cite{fpn} for fair comparisons, and all the backbones are pre-trained on ImageNet-1K.
We adopt AdamW \cite{adamw} optimizer with cosine learning rate schedule \cite{cosine},
while the initial learning rate is 1e-4.
The input images are randomly resized and cropped to 512$\times$512 for training, and the shorter sides of images are set to 512 while testing.

\noindent\textbf{Results.} 
The results on ADE20k dataset are shown in Table \ref{sota_seg}. Our MorphMLP outperforms ResNet\cite{regnet} and PVT\cite{pvt} significantly. Compared with Swin-T, our MorphMLP-T can achieve better mIoU  with fewer parameters (26.4M \textit{vs.}  31.9M). 

\subsection{Ablation Study}
For Table \ref{ablation_length}, \ref{ablation_spatial}, and \ref{ablation_importantce},
we train all the models based on MorphMLP-T for 100 epochs on ImageNet. To explore the variants of our spatial-temporal design,  we adopt {MorphMLP}-S as the backbone on SSV1.


\noindent
\textbf{Impact of chunk length.} 
In the MorphMLP, we expand the chunk length gradually.
The spatial resolutions of feature maps of Stages 1-4 are 56, 28, 14, 7, respectively. For the horizontal/vertical directions, chunk lengths 14, 28, 28, 49 in Stages1-4 can cover quarter, one, two and all rows/columns of the tokens, respectively, which can discover core semantics  progressively by operating the FC filter from small to big spatial region. 

As shown in Table \ref{ablation_length}, there are some alternative ways to set chunk length.  The first line represents that $\tt MorphFC_s$ in each stage covers one row of image/video tokens, which only models global information. The second, third  and fourth line utilize the small chunk length, which only captures local structure.  The results show that our progressively expanding  pattern can perform better than the solely local or global pattern.   The reason is that, in the shallow layer, the original texture and shape information of the image/videos is relatively intact. Therefore, it is critical to capture detailed structures in the early stage.  The features in the deep layers cover more semantic information, thus long-range relation modeling is significant. \textbf{Note} that the improvement brought by expanding chunk lengths on video is larger than image because such pattern is conducive to discovering more fine-grained semantics for many tiny movement actions.  

It is also worth noting that since chunk sizes are equal to H, W of features in
each stage if input is 224×224, 1st row is no
`Morph’ (progressively discovering core semantics), but with hierarchical downsampling only. Last row is our final model (w/ both
`Morph’ and same downsampling as the 1st row). Comparisons show benefits are from MorphFC design instead of hierarchical downsampling.

\begin{table}[t]
\centering
    \begin{minipage}[t]{0.547\linewidth}
    \caption{Impact of chunk length.}
        \centering
        \setlength{\tabcolsep}{2.0pt}
        \resizebox{1\textwidth}{!}{
        \begin{tabular}{cccc|ccc}
        \Xhline{0.6pt}
        \multirow{2}*{Stage1}& \multirow{2}*{Stage2} & \multirow{2}*{Stage3}  & \multirow{2}*{Stage4}  &SSV1 & ImageNet & ADE20K \\
        ~ & ~ & ~ & ~ &Top-1 & Top-1  & mIoU\\
        \Xhline{0.3pt}
        56  & 28  &14 &7 &48.6&79.1 & 41.6  \\
        3 & 3 & 3 &3 &48.0 &78.2 & 41.0 \\
        7 & 7 & 7 & 7 &48.4 &79.0& 41.9 \\
        14 & 14 & 14 & 14& 48.7 &79.0 & 42.0 \\
        14 & 28 & 28 &7 &49.2& 79.3& 42.0 \\
        28 & 28 & 28 &49 & 49.1& 79.4& 42.3 \\
	\rowcolor{gray!20}
        14 & 28 & 28 &49 &\textbf{50.6}&\textbf{79.6} &\textbf{42.6}\\
        \Xhline{0.6pt}
        \end{tabular}
        }
    \label{ablation_length}
    \end{minipage}
    \begin{minipage}[t]{0.43\linewidth}
    \caption{Detail designs of spatial-temporal MorphMLP block.}

        \centering
        \setlength{\tabcolsep}{2.4pt}
        \resizebox{1.0\textwidth}{!}{
            \begin{tabular}{cccc|c}
        \Xhline{0.6pt}
        \multirow{2}*{Method} & \multirow{2}*{Order} & Standard & Skip & SSV1 \\
        ~ & ~ & Residual & Residual & Top-1 \\
        \Xhline{0.3pt}
        Parallel & T$\|$S & \CheckmarkBold & ~ & 49.2 \\
        \Xhline{0.3pt}
        \multirow{4}*{Sequential} & T+S & \CheckmarkBold & ~ & 49.8  \\
        ~ & S+T & \CheckmarkBold & ~ & 50.2  \\
        ~ & \cellcolor{gray!20}{T+S} & \cellcolor{gray!20}{~} & \cellcolor{gray!20}{\CheckmarkBold} & \cellcolor{gray!20}{\textbf{50.6}}  \\
        ~ & S+T & ~ & \CheckmarkBold &  31.7 \\
        \Xhline{0.6pt}
        \end{tabular}
        }
    \label{ablation_residual}
    \end{minipage}

\end{table}

\noindent

\noindent
\textbf{Detail designs of spatial-temporal MorphMLP block.}
We explore some alternative designs for our spatial-temporal MorphMLP block in the Table \ref{ablation_residual}. To begin with, in addition to applying $\tt MorphFC_{t}$ and $\tt MorphFC_s$ in a sequential way, we can add the features from $\tt MorphFC_{t}$ and $\tt MorphFC_s$ in parallel. As shown in Table \ref{ablation_residual}, the parallel way performs worse than the sequential way. We argue that it is more difficult for joint spatial and temporal optimization. Moreover, we explore different spatial-temporal orders and residual connections. Standard residual refers to applying a residual connection after each module in MorphMLP block of Fig. \ref{fig:video_block}. Skip residual means that a connection is applied between input features of MorphMLP block and output features of the $\tt MorphFC_s$  (red line in Figure \ref{fig:video_block}).  The results show that sequential temporal and spatial order with skip residual connection is the optimal setting.

\noindent
\textbf{Comparisons with convolution.}
To compared with spatial convolution, we replace the $\tt MorphFC_s$ layer with typical 3$\times$3  and 7$\times$7 convolution on image domain. As shown in Table \ref{ablation_spatial}, our $\tt MorphFC_s$ can outperform typical convolution by a large margin. This demonstrates that typical convolution is difficult to capture long-range information, which is crucial to the recognition problem. Furthermore, we adopt two 1D group convolutions along the horizontal and vertical direction, whose kernel sizes are exactly the same as our chunk lengths in each stage. 
The results show that our method is much better than group conv in terms of speed and accuracy. This indicates the effectiveness of our $\tt MorphFC_s$. 

Moreover, we do comparisons between $\tt MorphFC_{t}$ and typical temporal convolutions, i.e., 3$\times$1$\times$1 and 5$\times$1$\times$1. As shown in Table \ref{ablation_temporal}, our $\tt MorphFC_{t}$ outperforms typical temporal convolutions greatly. This is because that typical convolutions only focus on local temporal information aggregation. On the contrary,  our  $\tt MorphFC_{t}$ is able to capture long-term temporal dependencies.
\begin{table}[t]
    
    \centering
    \begin{minipage}[t]{0.48\linewidth}
    \caption{Spatial design of MorphMLP.}

        \centering
        \setlength{\tabcolsep}{1.0pt}
        \resizebox{\linewidth}{!}{
        \begin{tabular}{c|ccc|c}
        \Xhline{0.6pt}
        \multirow{2}*{Operation} & \#Param. & FLOPs & Throughput & ImageNet \\
        ~ & (M) & (G) & (images/s) & Top-1 \\
        \Xhline{0.3pt}
        3$\times$3 Conv & 34.5 & 6.2 & 676 & 77.3  \\
        7$\times$7 Conv & 113 & 20.6 & 532 & 77.7  \\
        Group Conv & 23.4 & 4.0 & 620 & 79.0  \\
	\rowcolor{gray!20}
        $\tt  MorphFC_{s}$ & 23.4 & 4.0 & 734 & \textbf{79.6} \\
        \Xhline{0.6pt}
        \end{tabular}
    }
    \label{ablation_spatial}
    \end{minipage}
    \begin{minipage}[t]{0.48\linewidth}
    \caption{Different operations.}
     
        \centering
        \setlength{\tabcolsep}{1.0pt}
        \resizebox{\linewidth}{!}{
         \begin{tabular}{ccc|cc}
        \Xhline{0.6pt}
        \multirow{2}*{Dimension} & \multirow{2}*{Style} & \multirow{2}*{Weight Sum} &SSV1& ImageNet \\
        ~ & ~ & ~ &Top-1& Top-1 \\
        \Xhline{0.3pt}
	\rowcolor{gray!20}
        H+W+C & Transformer & \CheckmarkBold & 50.6&\textbf{79.6} \\
        H+W & Transformer & \CheckmarkBold &49.4& 78.5 \\
        H+W+C & CNN & \CheckmarkBold &47.2& 77.2 \\
        H+W+C & Transformer & \XSolidBrush &50.2& 79.3 \\
        \Xhline{0.6pt}
        \end{tabular}
        }
    \label{ablation_importantce}
    \end{minipage}

    \begin{minipage}[t]{0.49\linewidth}
    \caption{Temporal design.}
        
        \centering
        \setlength{\tabcolsep}{1pt}
        \resizebox{0.99\textwidth}{!}{
        \begin{tabular}{c|cc|c}
        \Xhline{0.8pt}
        Operation & \#Param.(M) & FLOPs(G)  & SSV1 \\
        \Xhline{0.3pt}
        3$\times$1$\times$1 Conv & 46.0  & 62.7& 47.9  \\
        5$\times$1$\times$1 Conv & 52.5 & 72.9  & 48.6  \\
	\rowcolor{gray!20}
        $\tt  MorphFC_{t}$ & 47.0 & 66.4  & \textbf{50.6}  \\
        \Xhline{0.8pt}
        \end{tabular}
    }
    \label{ablation_temporal}
    \end{minipage}
    \begin{minipage}[t]{0.48\linewidth}
    \caption{Training cost.}
       
        \centering
        \setlength{\tabcolsep}{1.5pt}
        \resizebox{\textwidth}{!}{
         \begin{tabular}{c|c|c|c|c}
            \Xhline{0.8pt}
            Video Model & TFLOPs  & K400 & Training & Cost \\
            \hline
            SlowFast &  1.11 & 71.0 & 30 epoch &444h\\
            Timesformer & 0.59 & 75.8 & 30 epoch &416h\\
	\rowcolor{gray!20}
            Morph-S  & 0.27 &77.0& 30 epoch & 408h\\
            \Xhline{0.8pt}
        \end{tabular}
        }
    \label{throughput}
    \end{minipage}

\end{table}

\noindent
\textbf{Importance of different operations.}
We explore the importance of different operations in Table \ref{ablation_importantce}. First, we evaluate the necessity of FC layers from three directions.
It shows that each direction plays an important role. 
Second, we replace the 3$\times$3 convolution with our $\tt MorphFC_s$ layer in ResNet\cite{resnet}/R(2+1)D\cite{r(2+1)d} and the result shows that Transformer structure is more suitable for our $\tt MorphFC$ than the bottleneck block of CNN.  Third, following the ViP\cite{vip}, we utilize a weighted sum after three directions FC layers. Results show that weighted sum can bring a slight improvement (0.3\%).

\noindent\textbf{Training speed.} As shown in Table \ref{throughput}, considering speed and accuracy trade-off, our approach is more efficient for training with other SOTA video methods.

\section{Conclusion}

In this paper, we propose a self-attention free, MLP-Like backbone for video representation learning, named MorphMLP. MorphMLP is capable of progressively discovering core semantics and capturing long-term temporal information.   To our best knowledge, we are the first to apply MLP-Like architecture in the video domain. The experiments demonstrate that  such self-attention free models can be as strong as  and even outperform self-attention based architectures.

\noindent\textbf{Acknowledgements.}
This project is supported by the National Research Foundation, Singapore under its NRFF Award NRF-NRFF13-2021-0008, and Mike Zheng Shou's Start-Up Grant from NUS. David Junhao Zhang is supported by NUS IDS-ISEP scholarship.


\clearpage
%
%
\bibliographystyle{splncs04}
\bibliography{morph}
\end{document}